\documentclass{article}

\usepackage[dblblindworkshop, final]{neurips_2025}
\workshoptitle{The Second Workshop on GenAI for Health: Potential, Trust, and Policy Compliance}

\usepackage[utf8]{inputenc} %
\usepackage[T1]{fontenc}    %
\usepackage{hyperref}       %
\usepackage{url}            %
\usepackage{booktabs}       %
\usepackage{amsfonts}       %
\usepackage{nicefrac}       %
\usepackage{microtype}      %
\usepackage{xcolor}         %
\usepackage{booktabs}
\usepackage{multirow}
\usepackage{xcolor}
\usepackage{colortbl}
\usepackage{longtable}
\usepackage{geometry}
\usepackage{graphicx}
\usepackage{amsmath}
\usepackage{amssymb}
\usepackage{enumitem}
\usepackage{tabularx}
\usepackage{textgreek}
\usepackage{wrapfig}
\title{Traj-CoA: Patient Trajectory Modeling via Chain-of-Agents for Lung Cancer Risk Prediction}

\author{%
  Sihang Zeng$^\text{\textalpha}$  \quad
  Yujuan Fu$^\text{\textalpha}$  \quad
  Sitong Zhou$^\text{\textalpha}$  \quad
  Zixuan Yu$^\text{\textalpha}$  \quad
  Lucas Jing Liu$^\text{\textbeta}$ \quad
  Jun Wen$^\text{\textgamma}$ \quad
  \\
  \textbf{Matthew Thompson}$^\text{\textdelta}$ \quad
   \textbf{Ruth Etzioni}$^\text{\textbeta}$ \quad
   \textbf{Meliha Yetisgen}$^\text{\textalpha}$ \quad
   \vspace{1mm}
  \\
$^\text{\textalpha}$ University of Washington \quad
$^\text{\textbeta}$ Fred Hutch Cancer Center \quad
$^\text{\textgamma}$ Harvard University \quad
$^\text{\textdelta}$ Google \quad
}

\begin{document}

\maketitle

\begin{abstract}
Large language models (LLMs) offer a generalizable approach for modeling patient trajectories, but suffer from the long and noisy nature of electronic health records (EHR) data in temporal reasoning. To address these challenges, we introduce Traj-CoA, a multi-agent system involving chain-of-agents for patient trajectory modeling. Traj-CoA employs a chain of worker agents to process EHR data in manageable chunks sequentially, distilling critical events into a shared long-term memory module, EHRMem, to reduce noise and preserve a comprehensive timeline. A final manager agent synthesizes the worker agents' summary and the extracted timeline in EHRMem to make predictions. In a zero-shot one-year lung cancer risk prediction task based on five-year EHR data, Traj-CoA outperforms baselines of four categories. Analysis reveals that Traj-CoA exhibits clinically aligned temporal reasoning, establishing it as a promisingly robust and generalizable approach for modeling complex patient trajectories. 
Implementation of Traj-CoA is available on \url{https://github.com/zengsihang/Traj-CoA}.

\end{abstract}

\section{Introduction}
Longitudinal Electronic Health Records (EHRs) provide rich, temporal data for modeling patient trajectories and predicting clinical outcomes~\citep{jensen2012mining}. Effective temporal reasoning is critical to unlocking this potential~\citep{cui2025timer}. For instance, tracking a lung nodule's evolution is key to diagnosing cancer~\citep{ocana2024growth}. While traditional approaches required complex feature engineering and task-specific models~\citep{silva2022modelling,pmlr-v225-burger23a,zeng2025trajsurvlearningcontinuouslatent}, modern Large Language Models (LLMs) promise a more generalizable, zero-shot paradigm for clinical prediction~\citep{sellergren2025medgemmatechnicalreport,singhal2025toward,zhang2024ultramedical}. However, the promise of LLMs is hindered by the unique challenges of EHR data: extremely long patient histories and inherent noisiness of the recorded clinical data~\citep{cui2025timer,kruse2025zeroshotlargelanguagemodels}.

Patient trajectories accumulate multimodal data over years, creating records that often exceed the context windows of LLMs~\citep{kruse2025zeroshotlargelanguagemodels}. Even models with large context windows are hampered by the "lost-in-the-middle" problem, where performance degrades on long inputs as they struggle to attend to the middle of a long input sequence~\citep{liu2023lost,cui2025timer}. Recent efforts to adapt LLMs for longitudinal EHR~\citep{cui2025timer,kruse2025zeroshotlargelanguagemodels} have shown that LLMs tend to fail on temporal reasoning over long EHR. While methods like temporal instruction tuning were proposed~\citep{cui2025timer}, these efforts have been largely confined to short EHR data or intensive care unit (ICU) data (<16k tokens). Temporal reasoning on very long EHR data over 32k or even 128k tokens remains an unclear challenge.

EHR data are inherently heterogeneous and primarily designed to support clinical care rather than research~\citep{kim2019evolving}. Consequently, they often contain noise arising from inconsistent formats, typographical errors, missing data, and irregular sampling. For many predictive tasks, only a small portion of a patient’s record is informative, while abundant irrelevant data can obscure key predictive signals. Early methods sought to address these challenges by enforcing standardized EHR formats~\citep{rajkomar2018scalable} or applying extensive feature engineering during preprocessing~\citep{silva2022modelling,pmlr-v225-burger23a}, which are difficult to generalize across diverse healthcare systems. Recently, LLMs offer a more flexible and broadly applicable framework for processing such data~\citep{zhu2024largelanguagemodelsinformation}. Yet, existing applications of LLMs in the EHR domain remain limited: many are restricted to single data modalities~\citep{zhu2024prompting}, while others struggle to capture temporal dependencies in long and complex patient histories~\citep{cui2025timer,kruse2025zeroshotlargelanguagemodels}. These gaps highlight the need for a mechanism to isolate relevant events scattered throughout patient histories.

Several strategies exist to manage long-context inputs for LLMs, including retrieval-augmented generation (RAG) and memory-based methods~\citep{liu2025comprehensivesurveylongcontext}. More advanced are agent-based approaches, which leverage the planning, memory, and reflection capabilities of LLMs to create autonomous agents~\citep{xi2023risepotentiallargelanguage,liu2025comprehensivesurveylongcontext}. Frameworks like the chain-of-agents (CoA), for instance, use multi-agent collaboration to enhance reasoning over long contexts~\citep{zhang2024chainagentslargelanguage}. Despite their success in the general domain, the application of these methods in healthcare remains limited, primarily confined to question-answering tasks~\citep{zhang2025leveraging,li2025agenthospitalsimulacrumhospital}. Consequently, patient trajectory modeling with LLMs, typically prediction tasks, over long and noisy EHR remains an underexplored practical challenge.

To overcome these challenges, we propose Traj-CoA, a novel framework for patient trajectory modeling. Traj-CoA employs a chain-of-agents architecture~\citep{zhang2024chainagentslargelanguage} with an external memory system to perform complex temporal reasoning on long patient histories. Our framework accepts a unified XML input that minimizes feature engineering and decomposes the longitudinal EHR into manageable time-aware chunks for more effective reasoning and summarization. These chunks are processed through a multi-agent workflow comprising specialized worker agents and a manager agent, which interact with a long-term memory module (EHRMem). We use lung cancer risk prediction as a use case to demonstrate the framework's capabilities, though it holds the promise to be a general-purpose solution for other longitudinal EHR tasks.
The main contributions of this work are:
\begin{itemize}[itemsep=0.5pt]
    \item We propose Traj-CoA, a novel chain-of-agents framework for temporal reasoning over long and noisy EHRs.
    \item To handle data noisiness, worker agents process unified XML inputs in sequential time-aware chunks, extracting salient signals for the task while removing localized noise.
    \item To manage long contexts, Traj-CoA leverages multi-agent communication and EHRMem for effective temporal reasoning with global context.
    \item Under a zero-shot setting in a lung cancer risk prediction task, Traj-CoA outperforms machine learning (ML), deep learning (DL), fine-tuned BERT, vanilla LLM, and RAG baselines, demonstrating its potential to become a simple yet powerful generalizable framework for patient trajectory modeling.
\end{itemize}

\begin{figure}[h]
    \centering
    \includegraphics[width=\linewidth]{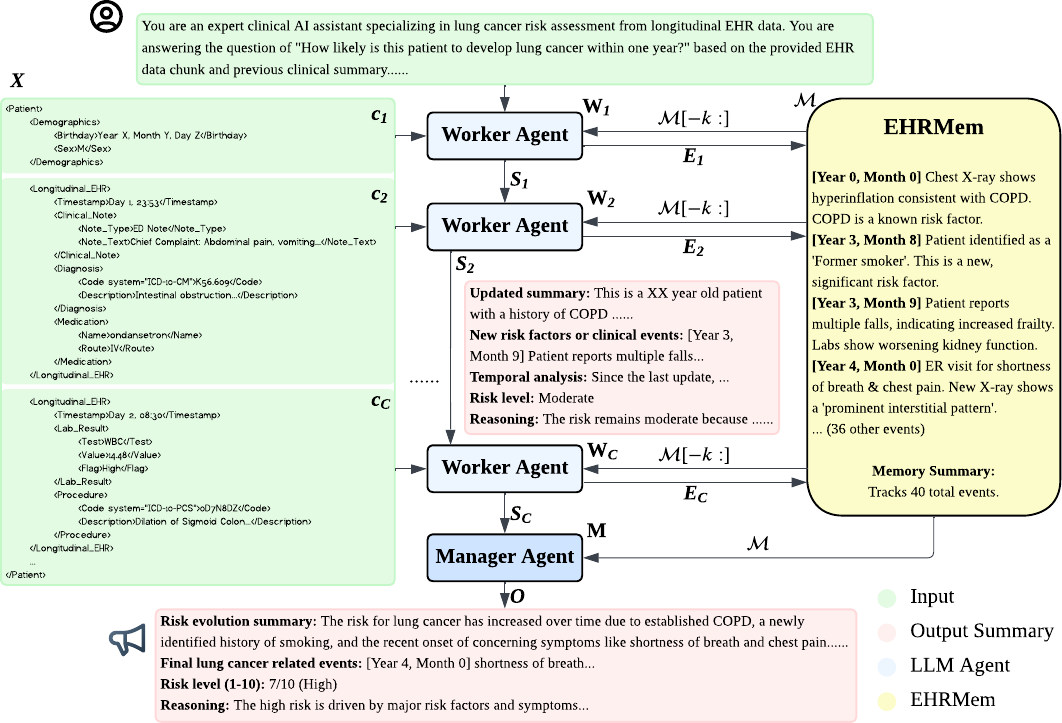}
    \caption{Traj-CoA architecture consisting of a chain of worker agents, a manager agent, and EHRMem. }
    \label{fig:architecture}
\end{figure}
\section{Related Works}
\paragraph{Patient Trajectory Modeling}
Extensive previous studies have explored patient trajectory modeling with longitudinal EHR. Conventional approaches rely on task-specific feature engineering for specialized models, including recurrent neural networks~\citep{silva2022modelling,pmlr-v225-burger23a,choi2017retaininterpretablepredictivemodel}, neural differential equations~\citep{zeng2025trajsurvlearningcontinuouslatent,moon2022survlatentodeneural,alaa2022ice}, and encoder-based transformers~\citep{li2020behrt,rasmy2021med,labach2023duettdualeventtime,placido2023deep}. 
More recently, EHR foundation models~\citep{renc2025foundationmodelelectronicmedical, wornow2025contextcluesevaluatinglong} have demonstrated zero-shot generalizability by pre-training on large structured datasets. However, they are often constrained by limited code sets and short context windows (<16k tokens), preventing them from fully leveraging the rich information in unstructured text or modeling very long patient histories.
LLMs-based approaches are a promising alternative, with demonstrated success in long-context reasoning across various domains~\citep{liu2025comprehensivesurveylongcontext}. However, applying a single, off-the-shelf LLM directly to model long patient trajectories has proven challenging. Recent studies show that even with access to a long context window, LLMs struggle with robust temporal reasoning in complex EHR data.
For instance, a longer context does not guarantee better temporal understanding, and RAG offers an incomplete solution~\citep{kruse2025zeroshotlargelanguagemodels}. Similarly, LLMs exhibit a "lost-in-the-middle" phenomenon on long EHRs, a problem partially mitigated by temporal instruction tuning~\citep{cui2025timer}. 
Therefore, a key open question remains: how can LLM-based approaches perform robust temporal reasoning on very long (over 32k or even 128k tokens) and noisy longitudinal EHR data without further training?

\paragraph{Long Context Modeling in LLMs}
While modern LLMs feature increasingly long context windows, their performance can degrade on lengthy inputs due to the "lost-in-the-middle" problem, where models overlook information positioned in the middle of the context~\citep{liu2023lost}. To address this limitation, several training-free strategies have emerged~\citep{liu2025comprehensivesurveylongcontext}. These include memory-based methods, which utilize an external memory to store and dynamically update the context~\citep{mem0,zhang2024surveymemorymechanismlarge}; RAG, which retrieves relevant information based on the query to augment the model's input~\citep{gao2024retrievalaugmentedgenerationlargelanguage}; and agent-based approaches. Agent-based systems leverage memory, reflection, planning, and inter-agent communication to reason over extended contexts~\citep{liu2025comprehensivesurveylongcontext}. For instance, frameworks like LongAgent and CoA segment the context into manageable chunks and employ multi-agent collaboration to process them~\citep{zhao2024longagentscalinglanguagemodels,zhang2024chainagentslargelanguage}. Although these techniques have been successfully applied to general-domain and medical question-answering tasks~\citep{zhang2025leveraging,li2025agenthospitalsimulacrumhospital,adams2025longhealth}, their utility for predictive tasks using longitudinal EHR is not well-established. In this study, we bridge this gap by proposing a novel method that synergizes memory and agent-based approaches for effective patient trajectory modeling.

\paragraph{Multi-Agent System for Biomedicine}
Multi-agent systems (MAS) in biomedicine employ multiple collaborating LLM agents to solve complex problems through role-playing and structured communication~\citep{wang2025surveyllmbasedagentsmedicine}. MAS has been successfully applied to biomedical discovery~\citep{qi2024largelanguagemodelsbiomedical}, diagnostics~\citep{chen2025enhancing}, and clinical trial optimization~\citep{yue2024clinicalagentclinicaltrialmultiagent}. For example, MedAgents~\citep{tang2024medagentslargelanguagemodels} assembles a multi-disciplinary team of specialized agents to improve zero-shot medical reasoning. Similarly, multiple specialized agents were orchestrated to emulate the tumor boards~\citep{MPH_2025}. 
While a recent study applied this collaborative paradigm to Alzheimer's prediction using longitudinal EHR~\citep{Li_Wang_Berlowitz_Mez_Lin_Yu_2025}, the effective application of MAS for \textit{temporal reasoning} over long and noisy EHR data remains unclear.
In contrast to prior work, Traj-CoA is a MAS for temporal reasoning in patient trajectory modeling.

\section{Method}
In this section, we introduce Traj-CoA, a generalizable framework for patient trajectory modeling with longitudinal EHR. We outline the problem formulation and Traj-CoA's core components: a unified data preprocessing pipeline that represents heterogeneous EHR in an XML format with time-aware chunking, a chain-of-agents (CoA) workflow that sequentially processes the long and noisy EHR, and a long-term EHR memory system that memorizes detailed clinical events. The framework is described in a general context before being applied to a specific use case of lung cancer risk prediction. An illustration of Traj-CoA is shown in Figure~\ref{fig:architecture}.

\subsection{Problem Formulation}
We consider a patient trajectory as a longitudinal, multimodal sequence of $n$ observations from the EHR, denoted as $\mathcal{X}=\{x_i, m_i, t_i\}_{i=1}^n$. Each tuple consists of a timestamp $t_i$, a data modality $m_i$ (e.g., diagnosis, lab result, clinical note), and the corresponding event data $x_i$ within the modality. A prediction task $\mathcal{T}$ is defined as a mapping $f:\mathcal{X}\rightarrow y$, where $y$ is the label of a task-specific outcome that occurs after the final observation time $t_n$. We seek a generalized framework that learns task $\mathcal{T}$ with minimum task-specific data preprocessing and model training.

The task is challenging due to the inherent properties of EHR data. The observation window ($t_1$ to $t_n$) may span over years. The data itself is heterogeneous, potentially from different EHR vendors and with inconsistent coding systems (e.g., ICD-9 vs. ICD-10) across patients; irregularly sampled, with non-uniform time gaps between observations; and noisy, containing missing data and irrelevant events to $\mathcal{T}$. An effective solution must therefore identify task-related events from this long, noisy data and model their temporal patterns in a generalizable manner. For instance, while lung nodule growth indicates cancer risk~\citep{ocana2024growth}, this temporal pattern is often scattered in EHRs, requiring algorithms to extract and interpret these critical dynamics.

In this study, we consider the lung cancer risk prediction task as a use case to evaluate the general Traj-CoA framework. Given the full patient trajectory $\mathcal{X}$ spanning up to 5 years before a reference date $t_n$, the task is to predict whether the patient will be diagnosed with lung cancer within the following year ($y=1$ for cases, $y=0$ for controls).

\subsection{Data Preprocessing}
To handle the noisy, long, and heterogeneous EHR data $\mathcal{X}$, we design a simple yet effective preprocessing pipeline to create a unified input representation $X$. Instead of relying on the complex task-specific feature engineering and cleaning common in prior work~\citep{pmlr-v225-burger23a,Wang_2020,wen2023lattelabelefficientincidentphenotyping}, our approach preserves the heterogeneity in the data while transforming it into an LLM-friendly format for reasoning. This consists of two steps: data unification into XML format and time-aware chunking.
\paragraph{Data Unification}

We convert a patient's entire multimodal history $\mathcal{X}$ into a single, unified XML format $X$. This strategy is motivated by the proven ability of LLMs to effectively comprehend structured, tag-based data~\citep{anthropic_xml_tags} and inspired by similar practice on EHR data~\citep{cui2025timer}. As illustrated in Figure \ref{fig:architecture}, we organize the longitudinal data chronologically within a nested XML structure. The root contains patient demographics, followed by a sequence of timestamped records, each encapsulating all data modalities and events observed at that time. This approach yields a clean, well-structured representation of the patient's timeline for Traj-CoA, preserving the data heterogeneity in text format.

\paragraph{Time-Aware Chunking}
Long XML inputs pose computational and reasoning challenges for LLMs. For example, in our lung cancer dataset, the 75th percentile token count reaches 120k (Table \ref{tab:dataset_stats}). Despite large context windows (>128k tokens) in recent LLMs, performance degrades significantly with longer contexts due to the "lost-in-the-middle" phenomenon, where models fail to process information from the middle sections effectively~\citep{liu2023lost}.

To address this limitation, we follow the CoA approach~\citep{zhang2024chainagentslargelanguage} which splits the context into chunks and uses multi-agent communication to ensure sequential information aggregation and seamless reasoning (see Section \ref{sec:coa}). Instead of hard chunking based on a fixed chunk size, we design a time-aware chunking strategy that avoids missing timestamp information caused by chunking. 
Specifically, we partition the XML EHR input into chunks of maximum $k$ tokens while preserving temporal ordering and timestamp completeness. Since our data structure is organized by timestamps, we split XML input into segments by timestamp, which are aggregated into chunks under the token limit $k$. When a single timestamp's records exceed $k$ tokens, we split them further while maintaining the original timestamp for each resulting chunk. This dynamic process converts the full XML input $X$ into a temporally coherent series of $C$ chunks $\{c_1, c_2..., c_C\}$, guaranteeing that all information within any given chunk is closely related in time. Note that the number of chunks $C$ may vary by patient.

\subsection{Chain-of-Agent}
\label{sec:coa}

We adapt the CoA algorithm~\citep{zhang2024chainagentslargelanguage} to reason over the chunked longitudinal EHR data. The vanilla CoA consists of two stages involving a chain of worker agents to conduct temporal reasoning chunk-by-chunk and a manager agent for the final prediction. Instead of relying on task-specific feature engineering or model training, CoA operates via task-specific instructions provided to its LLM agents, an approach that offers greater flexibility and efficiency for complex reasoning tasks.~\citep{zhang2024chainagentslargelanguage,sahoo2025systematicsurveypromptengineering} 

In stage one, a series of worker agents $\textbf{W}_i$ sequentially process each chunk $c_i$. Each worker agent takes the current chunk $c_i$, a task-specific instruction $I_W$, and the summary message $S_{i-1}$ from the preceding agent as input. Its function is to extract salient task-related information from $c_i$, analyze temporal patterns in relation to the aggregated summary, and produce an updated summary message $S_i$. This sequential process allows for the progressive task-related information aggregation across the entire longitudinal EHR. The operation of each worker agent is defined as:
\begin{align}
S_i = \textbf{W}_i \left(I_W, S_{i-1}, c_i\right)
\end{align}

In stage two, a manager agent $\textbf{M}$ receives the final summary message $S_C$ from the last worker agent $\textbf{W}_C$ along with a task-specific instruction $I_M$. The manager agent's role is to synthesize the comprehensive information contained in $S_C$ to produce the final output $O$. This is formulated as:
\begin{align}
O = \textbf{M}\left(I_M, S_C\right)
\end{align}

The CoA framework transforms a long-context reasoning problem into a structured agent communication chain, with each agent assigned a shorter context, thereby improving the reasoning quality and mitigating the LLM's "lost-in-the-middle" phenomenon common in long-context reasoning.~\citep{zhang2024chainagentslargelanguage}

\subsection{EHRMem}
In our experiments, we found that a direct application of the vanilla CoA framework to longitudinal EHR data can lead to the progressive abstraction and loss of critical task-related information over long sequences. In other words, early clinical events may be vital for accurate prediction but may be ``forgotten" by the final summary message $S_C$. To mitigate this, we introduce EHRMem, a structured long-term memory module storing task-related events and timestamps, denoted by $\mathcal{M}$.

EHRMem is populated during stage one, where each worker agent extracts new clinical events or risk factors that are potentially task-related, and stores their contents and timestamps as entries $E_i$ in $\mathcal{M}$. 
To prevent overwhelmingly redundant entries caused by EHR "copy-forwarding,"~\citep{wornow2025contextcluesevaluatinglong} we employ a deduplication mechanism: each agent's prompt is augmented with the last $k$ events from $\mathcal{M}$ and is instructed to only store new, unrecorded information.
In stage two, the manager agent's decision-making is augmented by the global context in $\mathcal{M}$, conditioning its output on both the final summary message $S_C$ and the entire memory $\mathcal{M}$. The Traj-CoA's operation is thus redefined as:
\begin{align}
S_i, E_i &= \textbf{W}_i \left(I_W, S_{i-1}, c_i, \mathcal{M}[-k:]\right)\\
\mathcal{M}&\leftarrow \mathcal{M}\oplus E_i\\
O &= \textbf{M}(I_M, S_C, \mathcal{M})
\end{align}
where $\mathcal{M}[-k:]$ denotes the last $k$ events in $\mathcal{M}$, and $\oplus$ means concatenation.
The EHRMem module serves two primary functions: (1) it constructs a distilled clinical timeline, effectively reducing the noise inherent in raw EHR, and (2) it provides the manager agent with a structured global context complementary to the unstructured worker agents' summary, enabling more robust reasoning across the entire patient history.

Crucially, the extraction heuristic for populating EHRMem is intentionally inclusive. Worker agents identify a slightly broader set of events \textit{potentially} relevant to $\mathcal{T}$, rather than \textit{strictly} filtering for those with an immediate, obvious connection to the task. This design choice acknowledges that local worker agents lack the global context to definitively assess an event's long-term significance. By preserving a richer, slightly redundant set of events in $\mathcal{M}$, we delegate the final synthesis and attribution of importance to the manager agent, which can leverage the complete temporal context for a more informed judgment.

\section{Experiments}

\subsection{Dataset}
We experimented on a proprietary case-control dataset on lung cancer risk assessment. Each instance in the dataset is anchored to a specific chest-related radiology exam (chest X-ray, chest CT, or abdomen CT) capable of visualizing the lungs. The prediction task is to determine, at the time of this index exam ($t_n$), whether a patient will receive a primary lung cancer diagnosis within 1 year. Cases are defined as subjects diagnosed with primary lung cancer within one year of the index exam, with diagnoses cross-validated against a state cancer registry. Controls are subjects with no cancer diagnosis. Cases and controls were matched at a 1:10 ratio on the time and type of the index exam. For each instance, up to five years of the patient's longitudinal EHR history prior to the index exam were recorded. This rich, multi-modal data encompasses both structured records (e.g., ICD codes, lab results, vitals) and unstructured text (e.g., clinical notes, radiology reports). 

We randomly sampled 13,629 instances (1,239 cases and 12,390 controls) for model development, which were further randomly split into a training (12,266 samples) and validation (1,363 samples) set. These data were used for fine-tuning and \textbf{not} used under a zero-shot setting. From the rest, we randomly sampled 300 instances (28 cases and 272 controls) to construct a test set for feasible and fair evaluation. We highlight that the token count of XML input is substantial, with an interquartile range (IQR) of 28k--121k for cases and 19k--132k for controls in the test set (Table~\ref{tab:dataset_stats}). A detailed data description is presented in the Appendix~\ref{app:data}.

\subsection{Experimental Settings}
We used MedGemma-27B~\citep{sellergren2025medgemmatechnicalreport} as the base model of Traj-CoA, with a default chunk size of 8k tokens and a maximum of 15 chunks to accommodate contexts up to 120k tokens. 
We benchmarked Traj-CoA against four baseline categories: (1) ML: logistic regression (LR) and XGBoost~\citep{chen2016xgboost} trained on summary EHR features. (2) DL: The RNN-based trajectory models RETAIN~\citep{choi2017retaininterpretablepredictivemodel} and PatientTM~\citep{silva2022modelling}. (3) BERT-based: Clinical ModernBERT (C-MBERT)~\citep{lee2025clinicalmodernbertefficientlong}, fine-tuned with LoRA~\citep{hu2021loralowrankadaptationlarge} using an 8k context window. (4) LLM: Zero-shot MedGemma using two strategies: direct prompting (Vanilla, up to 64k context) and RAG with a bge-m3~\citep{chen2024bgem3embeddingmultilingualmultifunctionality} retriever on the time-aware chunks.

For LR, XGBoost, and RETAIN, we used diagnosis codes as input features. For PatientTM, we used the text descriptions of medical codes and unstructured notes as input. For BERT and LLM methods, we used the same XML input and employed a middle truncation strategy for context beyond the window. Specifically, we alternately selected texts from the beginning and end of the patient record until the context window limit is met, maintaining their relative chronological order. Compared to left truncation, where the most recent records are used, this method retains a longer and more challenging temporal span for evaluating temporal reasoning. We further report the performance using left truncation for BERT and vanilla baselines in Appendix~\ref{app:fullres}, preliminary results of fine-tuning Traj-CoA in Appendix~\ref{app:rft}, a case study in Appendix~\ref{app:case}, and error analysis in Appendix~\ref{app:error}.

We evaluated all methods on AUROC and the best F1 score among all thresholds, with its corresponding precision and recall. 
We report AUPRC and discuss the differences in the Appendix~\ref{app:fullres}. For BERT, the risk score was derived from the output logit. For all LLM-based methods, we prompted the model to output a risk score between 1 and 10. Further experimental details are in the Appendix~\ref{app:exp}.

\begin{table}[ht]
\centering
\caption{Performance comparison of different models. \textbf{Bold} indicates the best AUROC and F1 among all models or zero-shot models. SFT means supervised fine-tuning. The average performance (mean$\pm$std) for Traj-CoA across 5 runs with different random seeds is reported.}
\label{tab:model_comparison}
\small
\resizebox{\textwidth}{!}{%
\begin{tabular}{llcccccc}
\toprule
\textbf{Model Family} & \textbf{Model} & \textbf{Prediction} & \textbf{Context}  & \textbf{AUROC} & \textbf{Precision} & \textbf{Recall} & \textbf{F1} \\
                      &                & \textbf{Method}     & \textbf{Window}  &                &                &                    &                             \\
\midrule
\multirow{2}{*}{ML} & LR & SFT & --- & 0.741 & 0.306 & 0.393 & 0.344 \\
\cmidrule{2-8}
 & XGBoost & SFT & --- & 0.763 & 0.367 & 0.393 & 0.379 \\
\midrule
\multirow{2}{*}{DL} & RETAIN & SFT & --- & 0.757 & 0.346 & 0.321 & 0.333 \\
\cmidrule{2-8}
 & PatientTM & SFT & --- & 0.730 & 0.361 & 0.464 & \textbf{0.406} \\
\midrule
BERT              & C-MBERT           & SFT                 & 8k               & 0.749                  & 0.367              & 0.393           & 0.379       \\
\midrule
\multirow{4}{*}{\begin{tabular}[c]{@{}l@{}}LLM\\(MedGemma)\end{tabular}} 
                      & Vanilla     & Zero-shot           & 32k              & 0.743                 & 0.345              & 0.357           & 0.351       \\
\cmidrule{2-8}
 & RAG & Zero-shot & 1k $\times$ 32 & 0.753 & 0.221 & 0.607 & 0.324 \\
 \cmidrule{2-8}
    & \textbf{Traj-CoA} (w/o EHRMem) & Zero-shot & 8k $\times$ 15 & 0.748 & 0.183 & 0.821 & 0.299 \\
\cmidrule{2-8}
    & \textbf{Traj-CoA}       & Zero-shot           & 8k $\times$ 15     & \textbf{0.766} $\pm$ 0.019 & 0.358 $\pm$ 0.057           & 0.436 $\pm$ 0.105  & \textbf{0.380} $\pm$ 0.018 \\

\bottomrule
\end{tabular}%
}
\end{table}

\subsection{Results}
Table \ref{tab:model_comparison} reports the performance of each method. 
Vanilla MedGemma with a 32k context window and RAG with top 32 chunks of maximum 1k tokens each achieves an AUROC of 0.74--0.75. 
However, expanding vanilla MedGemma's context to 64k degrades its performance to an AUROC of 0.714, suggesting difficulty in utilizing longer input sequences effectively (Table~\ref{tab:additionalres}). In contrast, Traj-CoA, with its 120k context and dedicated design, substantially outperforms all zero-shot baselines, achieving an AUROC of 0.766 and an F1 score of 0.380, outperforming most SFT baselines and comparable to the best. These results highlight Traj-CoA's superior capability to conduct temporal reasoning over very long EHRs, overcoming the limitations observed in standard long-context models.

\paragraph{Ablation Study}
To analyze the contribution of the EHRMem component, we performed an ablation study by removing it from our architecture. As detailed in Table \ref{tab:model_comparison}, this modification significantly impairs model performance, causing an absolute drop of 1.8\% in AUROC and 8.1\% in F1 score. This result underscores the importance of maintaining a detailed long-term memory of clinical events, as it provides predictive signals that are not fully captured by the evolving summary alone.

\section{Analysis}
To further probe the mechanisms by which Traj-CoA synthesizes temporal information from longitudinal EHR, we conducted additional analyses to answer five research questions. We fixed a random seed for the following analyses.

\subsection{Sensitivity Analysis}
\subsubsection*{Q1: How does the chunk size affect the performance?}
\begin{wrapfigure}{r}{0.4\textwidth}
  \centering
  \includegraphics[width=0.38\textwidth]{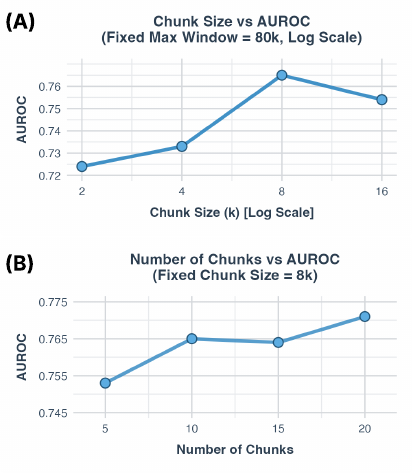}
  \caption{Sensitivity analysis on (A) chunk size and (B) number of chunks.}
  \label{fig:sensitivity}
  \vspace{-20pt} %
\end{wrapfigure}

Motivated by recent findings that LLM performance can degrade in very long contexts~\citep{liu2023lost}, we analyzed Traj-CoA's sensitivity to chunk size. To isolate this effect, we fixed the total context length at 80k tokens and varied the chunk size from 2k to 16k, which correspondingly adjusted the number of chunks from 40 down to 5.
Figure \ref{fig:sensitivity}A shows that performance peaks with a moderate chunk size of 8k, while both smaller (2k) and larger (16k) chunks result in lower AUROC scores.

This reveals a fundamental trade-off. Small chunks force a long chain of iterative summarizations, risking catastrophic forgetting~\citep{Kirkpatrick_2017} where early, critical details are abstracted away. Conversely, large chunks shorten the chain but are susceptible to the "lost-in-the-middle" issue~\citep{liu2023lost}, where each worker agent fails to identify fine-grained signals within a vast context. The 8k chunk size appears to provide an optimal balance, preserving local detail while enabling effective global aggregation.

\subsubsection*{Q2: How does the maximum context window affect the performance?}
We next analyzed how performance scales with the total context window to determine if more temporal information is beneficial. We fixed the chunk size at 8k and increased the maximum number of chunks from 5 to 20, thereby expanding the context window from 40k to 160k tokens.

As shown in Figure \ref{fig:sensitivity}B, AUROC consistently improves as the context window expands. While the vanilla LLM baseline also improves when scaling from an 8k to a 32k context, its performance degrades significantly at 64k (Table \ref{tab:additionalres}). In contrast, Traj-CoA maintains its positive trend up to 160k tokens. This demonstrates that EHRs contain rich, albeit noisy, predictive signals and that Traj-CoA's architecture is uniquely capable of leveraging these ultra-long sequences where standard methods fail.

\begin{figure}[ht]
    \centering
    \includegraphics[width=\linewidth]{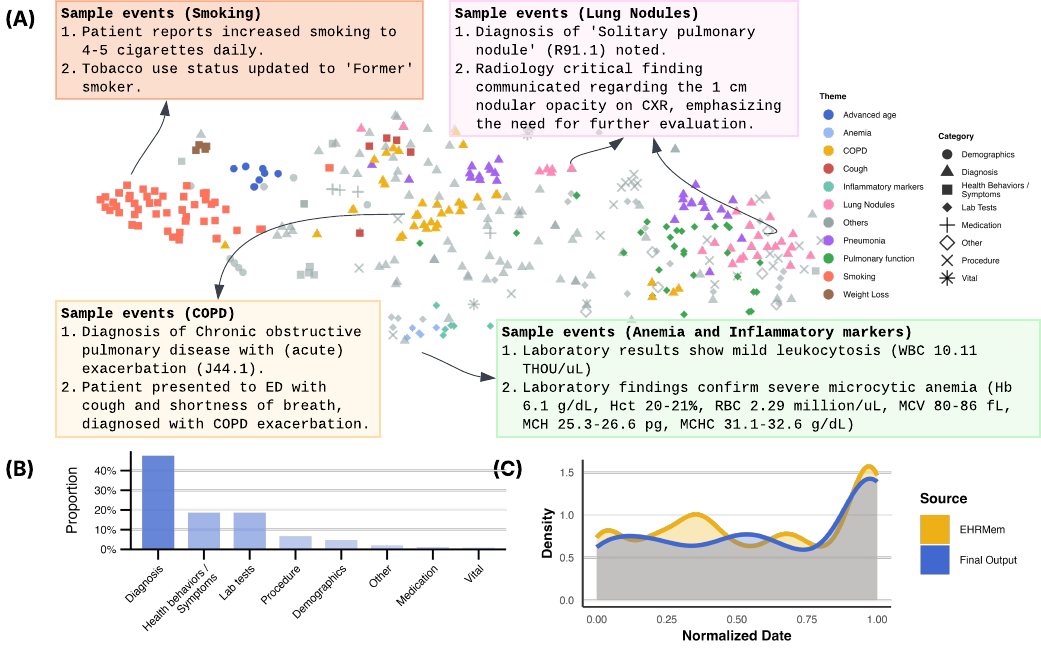}
    \caption{Analysis of Traj-CoA's behavior. (A) t-SNE plot visualizing the distribution of lung cancer related events in all cases' output $O$ and sample events (timestamps in sample events are omitted for de-identification purposes). The events were embedded through nomic-embed-text-v1.5~\citep{nussbaum2025nomicembedtrainingreproducible}; (B) Distribution of categories in the lung cancer related events; and (C) Normalized date distribution of the events.}
    \label{fig:temporal}
\end{figure}

\subsection{Temporal Reasoning Analysis}
To investigate Traj-CoA's temporal reasoning, we performed a topic modeling analysis on the clinical events it identified as salient for lung cancer prediction. We used an LLM-based pipeline inspired by TopicGPT~\citep{pham2024topicgptpromptbasedtopicmodeling} to systematically categorize these events.
First, for all positive cases, we extracted the relevant events from the model's final output $O$. Next, using Qwen2.5-7B-Instruct~\citep{qwen2025qwen25technicalreport}, we conducted a three-step analysis: (1) each event was classified into one of seven predefined categories (diagnosis, procedure, lab tests, vital, medication, health behaviors or symptoms, and demographics); (2) for each category, the top three common themes were generated based on all events under it; and (3) each event was mapped to its most relevant theme, if any, or recorded as ``Others". This structured thematic analysis of the model's output forms the basis to answering the subsequent Q3--Q5.

\subsubsection*{Q3: Does Traj-CoA reason across diverse event categories?}
We sought to verify that Traj-CoA's reasoning extends beyond trivial heuristics (e.g., identifying smoking status) to encompass a broad range of clinical events. Our analysis confirms that Traj-CoA identifies salient events from all seven predefined categories. As shown in the distribution in Figure \ref{fig:temporal}B, the most frequently utilized categories are diagnosis, health behaviors or symptoms, and lab tests.

The diversity of these identified events is further underscored by the t-SNE visualization in Figure \ref{fig:temporal}A. The event embeddings are scattered across the semantic space rather than forming a single cluster, indicating that the model draws upon a wide variety of clinical concepts. The qualitative examples presented in the figure corroborate this ability. This demonstrates that Traj-CoA performs multifaceted reasoning by integrating signals from a medically diverse set of events for its predictions.

\subsubsection*{Q4: Can Traj-CoA reason over the entire time horizon?}
To determine if Traj-CoA utilizes the full patient trajectory or suffers from the "lost-in-the-middle" phenomenon, we analyzed the temporal distribution of the events it identified as salient. In Figure \ref{fig:temporal}C, we plot the normalized date distribution of events from the EHRMem ($\mathcal{M}$) and the final output ($O$).

Both distributions show a concentration of events in the final year before the prediction date, which aligns with the clinical intuition that recent events are often most critical for diagnosis. Crucially, however, the model still identifies events from earlier periods. The presence of historical events in the final output demonstrates that Traj-CoA effectively synthesizes information over long time horizons, appropriately weighing recent events more heavily without discarding valuable historical context.

\subsubsection*{Q5: Is Traj-CoA's reasoning clinically relevant?}
Finally, we assessed the clinical relevance of Traj-CoA's reasoning by analyzing the themes of the salient events. The t-SNE visualization of the event embeddings (Figure \ref{fig:temporal}A) reveals that the most frequently identified themes form distinct semantic clusters, indicating thematic consistence. 

Importantly, these data-driven themes align directly with well-established clinical knowledge. The top themes identified include advanced age, anemia, COPD, cough, inflammatory markers, lung nodules, pneumonia, pulmonary function, smoking, and weight loss. These are either widely recognized by clinical practice and screening guidelines for assessing lung cancer risk~\citep{wolf2024screening,schabath2019cancer}, or consistent with existing evidence~\citep{kang2020clinical,engels2008inflammation,prado2023symptoms}. This validates the model's interpretability and demonstrates that its risk predictions are based on clinically meaningful patterns within the EHR data.

\subsection{Complexity Analysis}
\begin{wraptable}{r}{0.36\textwidth}
\caption{Time complexity.}
\label{tab:time_complexity_wrap}
\small
\begin{tabular}{lcc}
\hline
\textbf{Method} & \textbf{Encode} & \textbf{Decode} \\ \hline
Vanilla & $O(L^2)$ & $O(L L_O)$ \\
RAG & $O(L_R^2)$ & $O(L_R L_O)$ \\
Traj-CoA & $O(L L_C)$ & $O(L L_O)$ \\ \hline
\end{tabular}
\end{wraptable}

We compare the time complexity of Traj-CoA to vanilla prompting and RAG, similar to the previous study~\citep{zhang2024chainagentslargelanguage}. For each patient, let $L$, $L_C$, $L_R$, and $L_O$ denote the average lengths of the total input, a single chunk, the retrieved context for RAG, and each model output, respectively. 
As shown in Table~\ref{tab:time_complexity_wrap}, Traj-CoA has a lower encoding complexity than vanilla prompting but is more computationally intensive than RAG. This presents a trade-off between latency and contextual completeness. RAG achieves lower latency via selective retrieval at the risk of information loss, whereas Traj-CoA processes the entire context. The optimal choice depends on whether an application prioritizes real-time performance or comprehensive temporal reasoning over a patient's history.

\section{Discussion}
In this work, we introduced Traj-CoA, a framework designed to perform temporal reasoning on long and noisy longitudinal EHR data for patient trajectory modeling. 
By decomposing the reasoning process across agents that analyze moderate-sized data chunks, Traj-CoA effectively sidesteps the "lost-in-the-middle" problem, while a dedicated EHRMem module prevents forgetting of crucial early events. This design unlocks a key capability: unlike standard LLMs, Traj-CoA's performance on long EHR scales positively with context windows up to 160k tokens, effectively leveraging the rich predictive signals in complete patient trajectories.

While the positive results from this study shows promise for Traj-CoA, limitations exist. From a technical perspective, future efforts can enhance Traj-CoA's performance, including access to external knowledge~\citep{huang2025biomni,gao2025txagent}, using more powerful base models, and fine-tuning via multi-agent training approaches~\citep{marti2025,motwani2025maltimprovingreasoningmultiagent}. 
Moreover, while our analysis interprets Traj-CoA's temporal reasoning in terms of \textit{what} the salient events look like, further analysis is needed to understand \textit{how} these events are synthesized by the model for accurate prediction among different subpopulations.
Finally, although Traj-CoA is a task-agnostic framework, it requires carefully crafted prompts for specific tasks. Research into prompt optimization~\citep{ramnath2025systematicsurveyautomaticprompt} or data-driven hypothesis generation~\citep{qi2023largelanguagemodelszero,qi2024largelanguagemodelsbiomedical} could reduce this dependency and allow for more general modeling of patient trajectories. 
From a clinical application perspective, Traj-CoA was evaluated on a relatively small, single-institution cohort and one predictive task. We plan to conduct broader-scale validation to establish Traj-CoA's generalizability across diverse clinical settings and a wider array of prediction targets.

In conclusion, Traj-CoA provides a novel framework for patient trajectory modeling and bridges the gap between the generalist agentic AI~\citep{zhang2024ultramedical, huang2025biomni} and the temporal reasoning in longitudinal EHR. Traj-CoA demonstrates potential as a generalizable framework for patient trajectory modeling, though further design and validation are needed to make it more trustworthy.

\section*{Acknowledgements}
This work was supported by the National Institutes of Health (NIH)—National Cancer Institute (Grant Nos. 1R01CA248422-01A1).

\bibliography{main}
\bibliographystyle{unsrt}

\newpage
\appendix
\renewcommand{\thefigure}{S\arabic{figure}} %
\renewcommand{\thesubsection}{A.\arabic{subsection}}
\renewcommand{\thetable}{S\arabic{table}}
\renewcommand{\theHtable}{S\arabic{table}}%
\renewcommand{\theHfigure}{S\arabic{figure}}%

\section*{Appendix A. Additional Results}
\setcounter{figure}{0} %
\setcounter{table}{0}

\subsection{Dataset Description}
\label{app:data}
We present the dataset statistics of the test set in Table~\ref{tab:dataset_stats}. There are 28 cases and 272 controls in the dataset. The data is extracted from a in-house institutional medical center, whose EHR system has experienced a transition during the period. Therefore, the data has clinical notes from two EHR systems. The table summarizes features including patient demographics and metrics quantifying the volume of clinical data, such as the number of notes, diagnosis codes, and lab codes. All statistics are presented as median and interquartile range (IQR). Notably, both the case and control groups share a median record length or 'Year span' of 4.0 years, suggesting comparable observation periods. However, the groups differ in the volume of specific data types; for instance, cases have a higher median count of diagnosis codes (156.0 vs. 128.5), whereas controls have a higher median count of medication codes (64.0 vs. 42.0).

Notably, this dataset presents a challenging prediction problem because both the case and control cohorts are anchored to a radiology exam. Consequently, all patients in the dataset underwent radiological imaging for a clinical indication, requiring the model to distinguish radiological abnormalities associated with lung cancer from those attributable to other conditions, such as cardiovascular disease.
\begin{table}[htbp]
\centering
\caption{Dataset Statistics by Case-Control Status. Values shown as median (IQR) unless otherwise specified.}
\label{tab:dataset_stats}
\begin{tabular}{lcc}
\toprule
\textbf{Variable} & \textbf{Cases} & \textbf{Controls} \\
 & \textbf{(n=28)} & \textbf{(n=272)} \\
\midrule
Sex (Female/Male) & 14/14 & 115/157 \\
Year of last record & 2015.5 (2014.0--2018.0) & 2016.5 (2013.0--2018.0) \\
Year span & 4.0 (2.0--5.0) & 4.0 (1.0--5.0) \\
XML tokens & 61,270 (28,675--121,722) & 51,610 (19,442--132,777) \\
\# Timestamps & 42.5 (24.0--79.8) & 38.5 (14.0--96.0) \\
\# EHR System 1 notes & 7.5 (1.0--29.3) & 8.0 (0.0--29.0) \\ %
\# EHR System 2 notes & 3.5 (0.0--12.8) & 4.0 (0.0--14.0) \\ %
\# Radiology reports & 10.0 (4.8--17.3) & 8.0 (3.0--15.3) \\
\# Diagnosis codes & 156.0 (92.8--271.3) & 128.5 (42.5--349.3) \\
\# Medication codes & 42.0 (17.8--135.3) & 64.0 (9.0--184.5) \\
\# Procedure codes & 98.0 (51.5--197.8) & 106.5 (38.0--231.5) \\
\# Lab codes & 186.0 (26.8--439.3) & 201.0 (49.8--523.0) \\
\# Vital sign codes & 28.0 (0.0--91.0) & 20.0 (0.0--79.5) \\
\bottomrule
\end{tabular}
\end{table}

\subsection{Full Results}
\label{app:fullres}
Table~\ref{tab:additionalres} presents the complete results for our method, Traj-CoA, alongside BERT and LLM-based baselines. Traj-CoA, configured with a maximum chunk size of 8k, achieves the highest AUROC (0.753 -- 0.771), outperforming all BERT-based and LLM baselines.

While we observe a divergence between AUROC and AUPRC scores, we argue that AUROC is the more appropriate primary metric for this clinical prediction task. Recent work by McDermott et al.~\citep{McDermott_Zhang_Hansen_Angelotti_Gallifant_2024} demonstrates that while AUROC favors model improvements uniformly across all positive samples, AUPRC can be a misleading metric that disproportionately rewards improvements for samples assigned high scores. In the context of cancer risk prediction, the clinical cost of false negatives is exceptionally high, making the correct classification of lower-scoring, at-risk individuals paramount. Since AUPRC's prioritization runs counter to this clinical need~\citep{McDermott_Zhang_Hansen_Angelotti_Gallifant_2024}, we assert that Traj-CoA's superior performance on the more robust and relevant AUROC metric is the most significant finding.
\begin{table}[ht]
\centering
\caption{Full performance comparison of BERT-based and LLM baselines and Traj-CoA on the lung cancer risk prediction task using the left or middle truncation strategy.}
\label{tab:additionalres}
\small
\resizebox{\textwidth}{!}{%

\begin{tabular}{lllllrrrrr}
\toprule
\textbf{Model Family} & \textbf{Model} & \textbf{Prediction method} & \textbf{Context Window} & \textbf{Truncation} & \textbf{AUROC} & \textbf{AUPRC} & \textbf{Precision} & \textbf{Recall} & \textbf{F1} \\
\midrule
\multirow{2}{*}{Clinical ModernBERT} & \multirow{2}{*}{BERT} & \multirow{2}{*}{SFT} & 8k & Left & 0.734 & 0.256 & 0.323 & 0.357 & 0.339 \\
 &  &  & 8k & Middle & 0.749 & 0.310 & 0.367 & 0.393 & 0.379 \\
\midrule
\multirow{21}{*}{MedGemma} & \multirow{10}{*}{Vanilla} & \multirow{10}{*}{Zero-shot} & 4k & Left & 0.627 & 0.133 & 0.191 & 0.444 & 0.267 \\
 &  &  & 8k & Left & 0.668 & 0.211 & 0.370 & 0.357 & 0.364 \\
 &  &  & 16k & Left & 0.738 & 0.251 & 0.266 & 0.607 & 0.370 \\
 &  &  & 32k & Left & 0.739 & 0.249 & 0.313 & 0.357 & 0.333 \\
 &  &  & 64k & Left & 0.737 & 0.235 & 0.233 & 0.500 & 0.318 \\
 &  &  & 4k & Middle & 0.646 & 0.139 & 0.169 & 0.393 & 0.237 \\
 &  &  & 8k & Middle & 0.647 & 0.171 & 0.217 & 0.357 & 0.270 \\
 &  &  & 16k & Middle & 0.732 & 0.256 & 0.500 & 0.250 & 0.333 \\
 &  &  & 32k & Middle & 0.743 & 0.242 & 0.345 & 0.357 & 0.351 \\
 &  &  & 64k & Middle & 0.714 & 0.214 & 0.237 & 0.500 & 0.322 \\
\cmidrule(l){2-10}
 & \multirow{4}{*}{RAG} & \multirow{4}{*}{Zero-shot} & 1k $\times$ 32 & - & 0.753 & 0.208 & 0.221 & 0.607 & 0.324 \\
 & & & 2k $\times$ 16 & - & 0.740 & 0.224 & 0.313 & 0.357 & 0.333 \\
 &  &  & 4k $\times$ 8 & - & 0.731 & 0.190 & 0.247 & 0.643 & 0.356 \\
 &  &  & 8k $\times$ 4 & - & 0.735 & 0.179 & 0.215 & 0.500 & 0.301 \\
\cmidrule(l){2-10}
 & \multirow{7}{*}{Traj-CoA} & \multirow{7}{*}{Zero-shot} & 8k $\times$ 5 & Middle & 0.753 & 0.203 & 0.262 & 0.393 & 0.314 \\
 &  &  & 8k $\times$ 10 & Middle & 0.765 & 0.205 & 0.217 & 0.750 & 0.336 \\
 &  &  & 8k $\times$ 15 & Middle & 0.764 & 0.233 & 0.291 & 0.571 & 0.386 \\
 &  &  & 8k $\times$ 20 & Middle & 0.771 & 0.214 & 0.255 & 0.500 & 0.337 \\
 &  &  & 2k $\times$ 40 & Middle & 0.724 & 0.168 & 0.131 & 0.893 & 0.228 \\
 &  &  & 4k $\times$ 20 & Middle & 0.733 & 0.174 & 0.203 & 0.464 & 0.283 \\
 &  &  & 16k $\times$ 5 & Middle & 0.754 & 0.204 & 0.163 & 0.857 & 0.274 \\
\bottomrule
\end{tabular}
}
\end{table}

\subsection{Preliminary Results of Fine-tuning Traj-CoA}
\label{app:rft}
To investigate if further training could enhance the predictive performance of Traj-CoA, we conducted a preliminary study using rejection sampling fine-tuning (RFT)~\citep{yuan2023scalingrelationshiplearningmathematical}. While our initial results are promising, we caution that these findings are exploratory. A systematic application of this method would necessitate a rigorous experimental design and extensive hyperparameter tuning to ensure robust and generalizable improvements.
\paragraph{Training Data Generation}
We generated a high-quality dataset for RFT using a rejection sampling methodology, beginning with an initial pool of 500 cases and 500 controls from the training set. For each patient, we generated four candidate reasoning trajectories using Traj-CoA (8k $\times$ 15 chunks setting) with a high sampling temperature of 1.5. A trajectory is defined as the sequence of inputs and outputs from all worker and manager agents during a single forward pass.

To select for high-quality reasoning, we applied the following rejection criteria: for cases, we retained the trajectory that produced the highest predicted risk score, provided it was greater than 6; for controls, we retained the trajectory yielding the lowest score, provided it was less than 4. From each retained trajectory, we constructed RFT samples by compiling the input-output pairs from the first and last worker agents, two randomly sampled intermediate worker agents, and the manager agent. This process yielded a final RFT dataset comprising 2,223 instruction-tuning samples for fine-tuning both agent types.

\paragraph{RFT}
We fine-tuned the MedGemma-27B model using supervised fine-tuning (SFT). For memory efficiency, we employed QLoRA~\citep{dettmers2023qlora} with 4-bit quantization, implemented with the Unsloth~\citep{unsloth} and Huggingface TRL~\citep{vonwerra2022trl} libraries. The LoRA rank and $\alpha$ were both set to 32. The model was trained for 3 epochs on two A100 GPUs, using a per-device batch size of 2 and 8 gradient accumulation steps, resulting in an effective batch size of 32. Because the RFT dataset contained a mixture of data for both worker and manager agents, the single fine-tuned model serves as a unified base for both agent types in our framework.
\paragraph{Preliminary Results}
As shown in Table~\ref{tab:sft}, RFT leads to a notable performance gain for Traj-CoA. For example, when configured with a 16k $\times$ 5 context window, RFT improves the AUROC from 0.754 to 0.789. This result suggests that fine-tuning on data generated via rejection sampling is a promising direction for enhancing model performance.

However, we observed two key disparities in these preliminary results. First, the improvement in AUROC was not consistently accompanied by an increase in the F1 score. Second, the performance gains were more pronounced in the 16k $\times$ 5 setting, despite the RFT data being generated from the 8k $\times$ 15 setting. We hypothesize that these inconsistencies may be attributable to the ratio of instruction-tuning data between worker and manager agents in the RFT dataset.
\begin{table}[ht]
\centering
\caption{Preliminary results for training Traj-CoA.}
\label{tab:sft}
\small
\resizebox{\textwidth}{!}{%
\begin{tabular}{llcccccc}
\toprule
\textbf{Model Family} & \textbf{Model} & \textbf{Prediction} & \textbf{Context}  & \textbf{AUROC} & \textbf{Precision} & \textbf{Recall} & \textbf{F1} \\
                      &                & \textbf{Method}     & \textbf{Window}  &                &                &                    &                         \\
\midrule
\multirow{4}{*}{\begin{tabular}[c]{@{}l@{}}MedGemma\\27B\end{tabular}} 
& \textbf{Traj-CoA}       & Zero-shot           & 8k $\times$ 15     & 0.764 & 0.291           & 0.571  & 0.386 \\
\cmidrule{2-8}
& \textbf{Traj-CoA w/ RFT}       & RFT           & 8k $\times$ 15     & 0.775 & 0.262           & 0.571  & 0.360 \\
\cmidrule{2-8}
& \textbf{Traj-CoA}       & Zero-shot           & 16k $\times$ 5     & 0.754 & 0.221           & 0.607  & 0.324 \\
\cmidrule{2-8}
& \textbf{Traj-CoA w/ RFT}       & RFT           & 16k $\times$ 5     & 0.789 & 0.241           & 0.714  & 0.360 \\

\bottomrule
\end{tabular}%
}
\end{table}

\paragraph{Future Directions}
While our exploratory experiments are promising, this work highlights several avenues for future research. 
First, our current approach trains a single, unified base model on mixed data for both worker and manager agents. This co-training strategy may introduce biases and limit the potential for agent specialization. Future work could explore dedicated multi-agent fine-tuning strategies~\citep{marti2025} to train distinct models for each role, potentially enhancing the performance of the overall framework.
Second, we employed an offline rejection sampling method for fine-tuning. Adopting online learning paradigms, such as reinforcement learning~\citep{marti2025,zhang2025medrlvremergingmedicalreasoning}, could enable more dynamic policy improvements and potentially lead to more robust and capable agents.
Finally, we observed that model performance can be sensitive to numerous hyperparameters. These include the case-control balance of the training data, the sampling temperature used for trajectory generation, and the ratio of worker-to-manager data. A rigorous, systematic study is needed to investigate the impact of these parameters and establish a more principled approach to optimizing the training process.
\subsection{Case Study}
\label{app:case}
We present a deidentified case study of Traj-CoA's final output $O$ for a case patient. As shown in Table~\ref{tab:case}, the patient is an elderly female with multiple comorbidities (e.g., COPD, cognitive impairment) and established risk factors, including an advanced age and a history as a former smoker. She recently had an emergency room visit for shortness of breath and a chest X-ray with a "prominent interstitial pattern," both of which are non-specific and could be attributed to her existing conditions. Traj-CoA demonstrates its ability to synthesize this heterogeneous information effectively. It correctly identifies that the combination of long-term risk factors and persistent, concerning symptoms warrants a high-risk assessment (8/10), showcasing its capability to produce predictions grounded in a holistic view of the patient's trajectory.

\begin{table}[h!]
    \centering
    \caption{Case study of the final output $O$ of a case patient.}

    \includegraphics[clip,trim=1in 1in 1in 1in, width=1.00\textwidth]{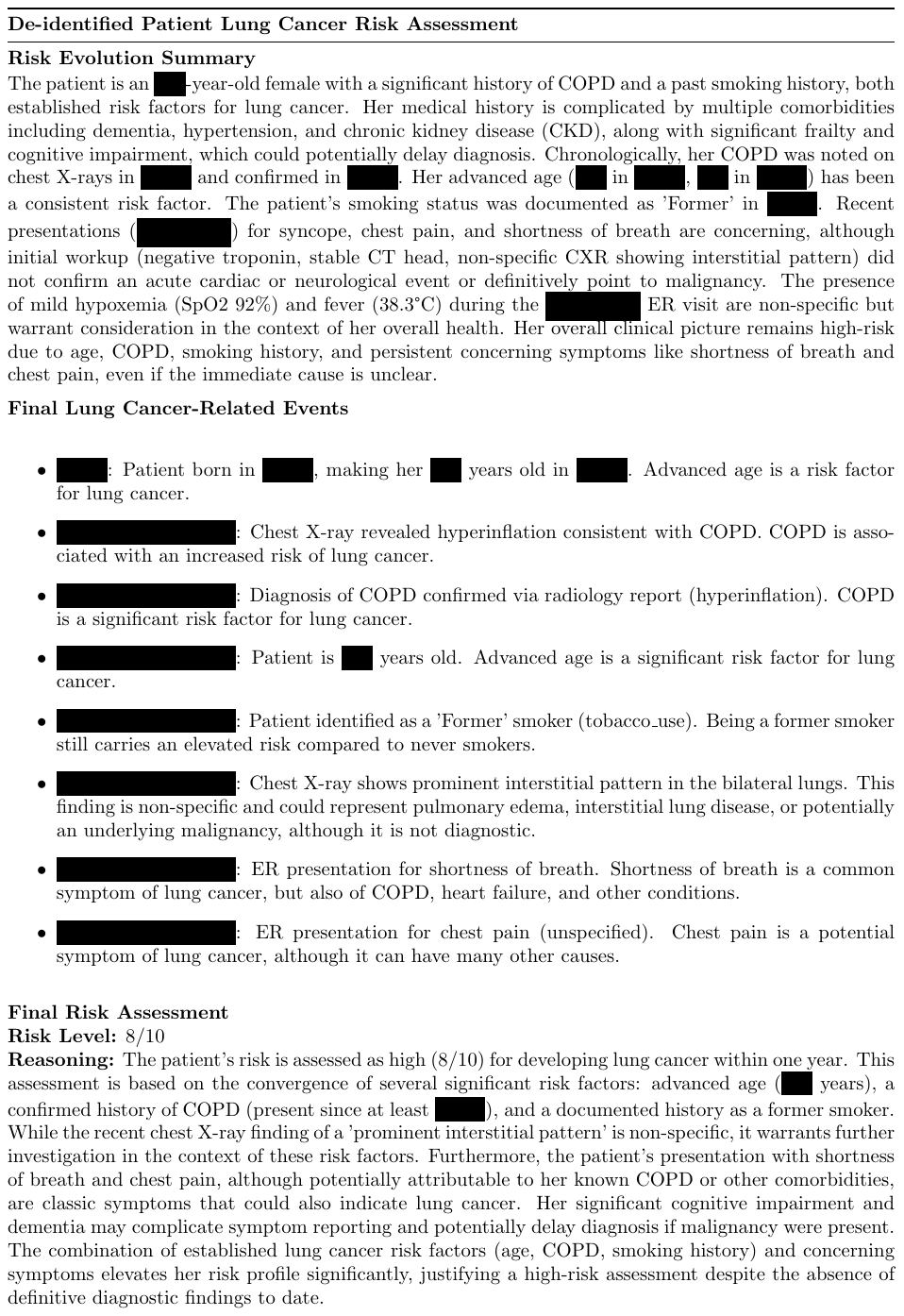}
    \label{tab:case}
\end{table}

\subsection{Error Analysis}
\label{app:error}

We conducted a qualitative analysis of the three \textbf{false-negative} cases, which were identified using a predicted risk threshold of 5.0. To investigate the model's failure modes, we performed a manual chart review of each patient's 5-year longitudinal EHR data preceding the index exam. This analysis of why Traj-CoA assigned erroneously low-risk scores aims to reveal its limitations and guide future improvements.

Case 1 is a patient in her early 70s with a complex medical history that includes a >40 pack-year smoking history, COPD, and pulmonary hypertension. While the patient presented with persistent cough, dyspnea, and chest pain, recent imaging showed no suspicious findings and indicated resolving pneumonia. Traj-CoA reasoned that these symptoms were manifestations of her known comorbidities, assigning a low predicted cancer risk of 3/10.

Case 2 is a patient in her 50s with a medical history notable for chronic cough and long-term immunosuppression due to arteritis. Although a PET-CT three years prior revealed a small nodule, the patient is a never-smoker with no family history of cancer, and subsequent CT scans showed no new suspicious nodules. Traj-CoA likely discounted the historical nodule due to its long-term stability and the patient's otherwise low-risk profile, assigning a low predicted cancer risk of 4/10.

Case 3 is a patient in her early 70s with a medical history that includes severe obesity, hypertension, and asthma. The patient reported a chronic cough and repeated exposure to fumes. She is a lifelong nonsmoker, and interim exams noted clear lungs. Traj-CoA correctly identified the exposure to fumes as a lung cancer related event, but attributed respiratory complaints to asthma and likely over-weighted the patient's nonsmoker status and negative exam results, thus assigning a low predicted cancer risk of 2/10.

In conclusion, our error analysis identifies two key limitations of the proposed model. First, as shown in all three cases, Traj-CoA demonstrates a \textbf{tendency to over-rely on recent, benign imaging results}, which can lead to poorly calibrated risk assessments that fail to accurately capture a patient's true risk. Second, as shown in cases 2 and 3, Traj-CoA may \textbf{underestimation of risk within the never-smoker subpopulation}, suggesting the model does not adequately capture the distinct predictive factors relevant to this group (e.g., environmental or occupational exposures)~\citep{cheng2021lung}.

These findings highlight opportunities for future research. To address miscalibration, future methods may incorporate methodologies designed to produce more reliable predictions, such as model fine-tuning. To improve fairness and accuracy, we advocate for the development of models that explicitly stratify by smoking status, either through cohort-specific architectures or by integrating domain knowledge of non-smoking-related risk factors.

\section*{Appendix B. Experimental Details}
\renewcommand{\thesubsection}{B.\arabic{subsection}}
\setcounter{subsection}{0}
\subsection{Experimental Settings}
\label{app:exp}
Our experiments were conducted using Python 3.10, Huggingface Transformers 4.53.2~\citep{wolf-etal-2020-transformers}, and vLLM 0.9.2~\citep{kwon2023efficient}. Models were downloaded from Huggingface\footnote{https://huggingface.co/Simonlee711/Clinical\_ModernBERT}\footnote{https://huggingface.co/google/medgemma-27b-text-it}. The BERT baseline was trained on a single NVIDIA A100 GPU; we added a linear layer over the [CLS] token's final hidden state to produce logits and optimized the model using a binary cross-entropy loss with a learning rate of 5e-5, a batch size of 8, and 4 gradient accumulation steps (effective batch size of 32). We employed an early stopping strategy with a patience of 3 epochs to prevent overfitting. For all LLM-based methods, we used default decoding parameters of Gemma 3: a temperature of 1.0, top-p of 0.95, and top-k of 64. To accommodate the model's size and the long-context requirements of the task, inference was performed on two NVIDIA A100 GPUs, leveraging tensor parallelism. 

\begin{table*}[ht]
\centering
\caption{Prompt for the initial worker agent.}

\small
\begin{tabularx}{\textwidth}{X}
\toprule
\underline{\textbf{\textsc{Initial Worker System Prompt}}} \\
\midrule
You are an expert clinical AI assistant specializing in lung cancer risk assessment from longitudinal EHR data. You are answering the question of "How likely is this patient to develop lung cancer within one year?" based on the provided EHR data chunk. \\
\\
\textbf{Task:} Analyze the first chunk of a patient's longitudinal EHR data, provided in XML format. Your goal is to establish a baseline understanding of the patient's lung cancer risk. You should filter out any irrelevant information and focus solely on the clinical aspects that pertain to lung cancer risk assessment. \\
\\
\textbf{Input:} \\
- \texttt{chunk\_xml}: A string containing the first segment of the patient's EHR data. \\
\\
\textbf{Instructions:} \\
1.  \textbf{Summarize the Clinical Information:} Briefly summarize the key clinical information present in this data chunk. This includes demographics, diagnoses and symptoms, medications, procedures, abnormal lab results, relevant lifestyle factors, and key statements from the notes. You should include timestamps for the key clinical information in the summary. Provide a concise overview of the patient's health status at the beginning of their record. \\
2.  \textbf{Identify Initial Risk Factors or Clinical Events:} Explicitly list all potential lung cancer risk factors or clinical events found in the data, such as risk factors, symptoms, abnormal lab results, findings, etc. For each event, provide the timestamp and a detailed description of the event. \\
3.  \textbf{Assess Initial Lung Cancer Risk:} Based on the identified lung cancer related risk factors or clinical events, provide an initial lung cancer risk assessment for this patient. The risk should be categorized as \textbf{Low}, \textbf{Moderate}, or \textbf{High}. Provide a clear rationale for your assessment. \\
\\
\textbf{Output Format:} \\
Your output must be a single, easily parsable JSON object with the following keys: \\
- \texttt{summary}: A string containing the clinical summary. \\
- \texttt{risk\_factors\_or\_clinical\_events}: A list of JSON objects, where each object details an identified lung cancer related risk factor or clinical event. \\
    \quad - \texttt{timestamp}: The timestamp of the event. \\
    \quad - \texttt{event}: A detailed description of the event. \\
- \texttt{risk\_assessment}: A JSON object indicating the assessed risk level for lung cancer diagnosis within 1 year ('Low', 'Moderate', or 'High'). \\
    \quad - \texttt{risk\_level}: The assessed risk level for lung cancer diagnosis within 1 year ('Low', 'Moderate', or 'High'). \\
    \quad - \texttt{reasoning}: A string explaining the basis for your risk assessment. \\
\\
ONLY output the JSON object without any additional text or formatting. Ensure that the JSON is valid and can be parsed easily. \\
\bottomrule
\end{tabularx}
\label{tab:prompt_initial_worker_system}
\end{table*}

\begin{table*}[ht]
\centering
\caption{User prompt for the initial worker agent.}

\small
\begin{tabularx}{\textwidth}{X}
\toprule
\underline{\textbf{\textsc{Initial Worker User Prompt}}} \\
\midrule
Here is the first data chunk: \\
\texttt{<chunk\_xml>} \\
\texttt{\{chunk\_1\_xml\}} \\
\texttt{</chunk\_xml>} \\
\\
Please provide the initial clinical summary and lung cancer risk assessment in JSON format. \\
\bottomrule
\end{tabularx}
\label{tab:prompt_initial_worker_user}
\end{table*}

\begin{table*}[ht]
\centering
\caption{Prompt for the subsequent worker agent.}

\small
\begin{tabularx}{\textwidth}{X}
\toprule
\underline{\textbf{\textsc{Subsequent Worker System Prompt}}} \\
\midrule
You are an expert clinical AI assistant specializing in lung cancer risk assessment from longitudinal EHR data. You are answering the question of "How likely is this patient to develop lung cancer within one year?" based on the provided EHR data chunk and previous clinical summary. \\
\\
\textbf{Task:} Analyze a new chunk of a patient's EHR data, considering the previous clinical summary, risk assessment, and the universal memory of lung cancer related events. Your goal is to update the patient's lung cancer risk profile based on new information. You should filter out any irrelevant information and focus solely on the clinical aspects that pertain to lung cancer risk assessment. \\
\\
\textbf{Input:} \\
- \texttt{previous\_summary}: A JSON object from the previous agent containing the summary, lung cancer related events, and risk assessment up to this point. \\
- \texttt{memory\_events}: A list of the last 10 lung cancer related events from the universal memory, providing historical context across all processed chunks. \\
- \texttt{new\_chunk\_xml}: A string containing the next segment of the patient's EHR data. \\
\\
\textbf{Instructions:} \\
1.  \textbf{Update the Summary:} Briefly summarize the key clinical information from the new data chunk and DO aggregate it with the previous summary. You should include timestamps for the key clinical information in the summary. Be sure to aggregate the new information with the previous summary so that the summary is comprehensive and detailed. Include all important timestamps so far. \\
2.  \textbf{Identify Risk Factors or Clinical Events:} List any new lung cancer risk factors or clinical events, such as risk factors, symptoms, abnormal lab results, findings, etc. \\
3.  \textbf{Analyze Temporal Patterns and Status Changes:} Describe any significant clinical changes or temporal trends observed between the previous data and this new chunk (e.g., progression of a disease, initiation of a new treatment). \\
4.  \textbf{Assess Updated Lung Cancer Risk:} Provide an updated lung cancer risk assessment, categorized as \textbf{Low}, \textbf{Moderate}, or \textbf{High}. Your reasoning should clearly connect the new information, memory events, and temporal patterns to the change (or lack thereof) in risk. \\
\\
\textbf{Output Format:} \\
Your output must be a single, easily parsable JSON object with the following keys: \\
- \texttt{updated\_summary}: A string with the summary of the entire clinical information so far. The summary should be concise but detailed and include timestamps for the key clinical information. \\
- \texttt{new\_risk\_factors\_or\_clinical\_events}: A list of JSON objects detailing the new lung cancer risk factors or clinical events that are NOT in the memory. Be comprehensive and detailed in the list of new events. \\
    \quad - \texttt{timestamp}: The timestamp of the event. \\
    \quad - \texttt{event}: A detailed description of the event, and how it may be related to lung cancer (risk factors, symptoms, abnormal lab results, findings, etc.) \\
- \texttt{temporal\_analysis}: A string describing clinical changes and temporal patterns so far. \\
- \texttt{updated\_risk\_assessment}: A JSON object for the updated risk level for lung cancer diagnosis within 1 year ('Low', 'Moderate', or 'High'). \\
    \quad - \texttt{risk\_level}: The updated risk level for lung cancer diagnosis within 1 year ('Low', 'Moderate', or 'High'). \\
    \quad - \texttt{reasoning}: A string explaining the rationale for the updated risk assessment. \\
\\
ONLY output the JSON object without any additional text or formatting. Ensure that the JSON is valid and can be parsed easily. \\
\bottomrule
\end{tabularx}
\label{tab:prompt_subsequent_worker_system}
\end{table*}

\begin{table*}[ht]
\centering
\caption{User prompt for the subsequent worker agent.}

\small
\begin{tabularx}{\textwidth}{X}
\toprule
\underline{\textbf{\textsc{Subsequent Worker User Prompt}}} \\
\midrule
Previous Agent Output: \\
\texttt{<previous\_summary>} \\
\texttt{\{previous\_agent\_output\}} \\
\texttt{</previous\_summary>} \\
\\
Memory Events (Last 10 from Universal Memory): \\
\texttt{<memory\_events>} \\
\texttt{\{memory\_events\}} \\
\texttt{</memory\_events>} \\
\\
New Data Chunk: \\
\texttt{<new\_chunk\_xml>} \\
\texttt{\{new\_chunk\_xml\}} \\
\texttt{</new\_chunk\_xml>} \\
\\
Please provide the updated and consolidated summary in JSON format. \\
\bottomrule
\end{tabularx}
\label{tab:prompt_subsequent_worker_user}
\end{table*}

\begin{table*}[ht]
\centering
\caption{Prompt for the manager agent.}

\small
\begin{tabularx}{\textwidth}{X}
\toprule
\underline{\textbf{\textsc{Manager Agent System Prompt}}} \\
\midrule
You are a senior clinical AI expert specializing in longitudinal lung cancer risk analysis. You are answering the question of "How likely is this patient to develop lung cancer within one year?" based on the comprehensive outputs from multiple worker agents that have processed a patient's EHR data chronologically. \\
\\
\textbf{Task:} Synthesize the outputs from the last worker agent and the universal memory of all lung cancer related events to provide a final, comprehensive lung cancer risk assessment and a narrative of the patient's risk evolution. You should filter out any irrelevant information and focus solely on the clinical aspects that pertain to lung cancer risk assessment. \\
\\
\textbf{Input:} \\
- \texttt{final\_worker\_outputs}: A JSON object, which is the output from the last worker agent that has processed a patient's EHR data chronologically. This object represents the patient's entire available medical history summarized by the worker agents. \\
- \texttt{universal\_memory\_events}: A list of all lung cancer related events from the universal memory, providing complete historical context across all processed chunks. \\
\\
\textbf{Instructions:} \\
1.  \textbf{Synthesize Temporal Trends:} Review the sequence of outputs and the complete universal memory. Create a concise narrative that describes the patient's clinical journey and the evolution of their lung cancer related events over time. Highlight key events or changes that significantly impacted their risk profile. \\
2.  \textbf{Final Lung Cancer Related Events Assessment:} Consolidate all identified lung cancer related events from the universal memory and worker outputs into a final, comprehensive list. Ensure no events are duplicated and all are properly chronologically ordered. \\
3.  \textbf{Assess Final Lung Cancer Risk:} Provide a final lung cancer risk assessment, from 1 to 10, where 1 is the lowest risk and 10 is the highest risk. \\
4.  \textbf{Provide Comprehensive Reasoning:} Justify your final risk assessment by explaining how the interplay of all lung cancer related events from the universal memory and their temporal evolution contributes to the patient's overall risk. This should be your most detailed and conclusive reasoning. \\
\\
\textbf{Output Format:} \\
Your output must be a single, easily parsable JSON object with the following keys: \\
- \texttt{risk\_evolution\_summary}: A string containing the narrative of the patient's clinical journey and risk evolution. \\
- \texttt{final\_lung\_cancer\_related\_events}: A list of strings containing all unique, consolidated lung cancer related events from the universal memory. \\
- \texttt{final\_risk\_assessment}: A JSON object for the final risk level for lung cancer diagnosis within 1 year (1 to 10, where 1 is the lowest risk and 10 is the highest risk). \\
    \quad - \texttt{risk\_level}: An integer from 1 to 10, where 1 is the lowest risk and 10 is the highest risk. \\
    \quad - \texttt{reasoning}: A string providing a comprehensive justification for the final risk assessment. \\
\\
ONLY output the JSON object without any additional text or formatting. Ensure that the JSON is valid and can be parsed easily. \\
\bottomrule
\end{tabularx}
\label{tab:prompt_aggregation_agent_system}
\end{table*}

\begin{table*}[ht]
\centering
\caption{User prompt for the manager agent.}

\small
\begin{tabularx}{\textwidth}{X}
\toprule
\underline{\textbf{\textsc{Manager Agent User Prompt}}} \\
\midrule
All Worker Agent Outputs: \\
\texttt{<final\_worker\_outputs>} \\
\texttt{\{final\_worker\_outputs\}} \\
\texttt{</final\_worker\_outputs>} \\
\\
Universal Memory Events (All Events): \\
\texttt{<universal\_memory\_events>} \\
\texttt{\{universal\_memory\_events\}} \\
\texttt{</universal\_memory\_events>} \\
\\
Please provide the final risk assessment and narrative summary in JSON format. \\
\bottomrule
\end{tabularx}
\label{tab:prompt_aggregation_agent_user}
\end{table*}

\begin{table*}[ht]
\centering
\caption{User prompt for the Manager agent without universal memory.}

\small
\begin{tabularx}{\textwidth}{X}
\toprule
\underline{\textbf{\textsc{Manager Agent User Prompt Without Memory}}} \\
\midrule
All Worker Agent Outputs: \\
\texttt{<final\_worker\_outputs>} \\
\texttt{\{final\_worker\_outputs\}} \\
\texttt{</final\_worker\_outputs>} \\
\\
Please provide the final risk assessment and narrative summary in JSON format. \\
\bottomrule
\end{tabularx}
\label{tab:prompt_aggregation_agent_user_no_mem}
\end{table*}

\begin{table*}[ht]
\centering
\caption{The query for RAG.}

\small
\begin{tabularx}{\textwidth}{X}
\toprule
\underline{\textbf{\textsc{Lung Cancer Query for RAG}}} \\
\midrule
What is the patient's risk of lung cancer?\\
\bottomrule
\end{tabularx}
\label{tab:lung_cancer_query}
\end{table*}

\section*{Appendix C. Prompts}
\renewcommand{\thesubsection}{C.\arabic{subsection}}
\setcounter{subsection}{0}
We present the prompt templates and query for RAG in Table \ref{tab:prompt_initial_worker_system}, \ref{tab:prompt_initial_worker_user}, \ref{tab:prompt_subsequent_worker_system}, \ref{tab:prompt_subsequent_worker_user}, \ref{tab:prompt_aggregation_agent_system}, \ref{tab:prompt_aggregation_agent_user}, \ref{tab:prompt_aggregation_agent_user_no_mem}, and \ref{tab:lung_cancer_query}.

\end{document}